\pdfoutput=1

\documentclass[11pt]{article}

\usepackage{acl}

\usepackage{times}
\usepackage{latexsym}

\usepackage[T1]{fontenc}

\usepackage[utf8]{inputenc}

\usepackage{microtype}
\usepackage{graphicx}
\usepackage{amsmath}
\usepackage{array}

\newcolumntype{L}[1]{>{\raggedright\let\newline\\\arraybackslash\hspace{0pt}}m{#1}}
\newcolumntype{C}[1]{>{\centering\let\newline\\\arraybackslash\hspace{0pt}}m{#1}}
\newcolumntype{R}[1]{>{\raggedleft\let\newline\\\arraybackslash\hspace{0pt}}m{#1}}

\usepackage{placeins}
\usepackage{float}
\usepackage{xcolor}

\title{Find The Gap: Knowledge Base Reasoning For Visual Question Answering}

\author{Elham J. Barezi \\
  Michigan State University \\
  \texttt{jebalbar@msu.edu} \\\And
  Parisa Kordjamshidi \\
  Michigan State University \\
  \texttt{kordjams@msu.edu} \\}

\begin{document}
\maketitle

\begin{abstract}
We analyze knowledge-based visual question answering, for which given a question, the models need to ground it into the visual modality and retrieve the relevant knowledge from a given large knowledge base (KB) to be able to answer. Our analysis has two folds, one based on designing neural architectures and training them from scratch, and another based on large pre-trained language models (LLMs). Our research questions are: 1) Can we effectively augment models by explicit supervised retrieval of the relevant KB information to solve the KB-VQA problem? 2) How do task-specific and LLM-based models perform in the integration of visual and external knowledge, and multi-hop reasoning over both sources of information? 
3) Is the implicit knowledge of LLMs sufficient for KB-VQA and to what extent it can replace the explicit KB? 
Our results demonstrate the positive impact of empowering task-specific and LLM models with supervised external and visual knowledge retrieval models. Our findings show that though LLMs are stronger in 1-hop reasoning, they suffer in 2-hop reasoning in comparison with our fine-tuned NN model even if the relevant information from both modalities is available to the model. 
Moreover, we observed that LLM models outperform the NN model for KB-related questions which confirms the effectiveness of implicit knowledge in LLMs however, they do not alleviate the need for external KB. 
\end{abstract}

\section{Introduction}

Visual Question Answering (VQA) aims to answer a question in natural language form while visual information is needed to provide the correct answer. Though over the past years, the progress in this field has been remarkable, existing models still suffer in answering the questions requiring knowledge beyond the image. These models suffer from lacking extra knowledge integration or integrating noisy knowledge in their reasoning procedure even when the source of required knowledge is given. This problem is formulated as the so-called KB-VQA where in addition to the question and image, the source of external knowledge is given as input. 

There have been various efforts to include external knowledge in the VQA task. These methods have two basic steps, retrieving related knowledge entries from the given knowledge bases, e.g., Wikipedia and ConceptNet~\cite{liu2004conceptnet}, and predicting the final answer by integrating this extracted external information, and question, and visual data. However, these retrieval-based approaches suffer from two weaknesses: (i) They use some heuristics like keyword matching for retrieval, and there is no guarantee for retrieving semantically relevant knowledge and (ii) the knowledge graphs are very dense, and plenty of irrelevant knowledge is retrieved even if the required knowledge is retrieved. These irrelevant entries mislead the reasoning procedure.
Since current information retrieval methods are based on simple heuristics and there is no guarantee for their retrieval accuracy,  they hurt the final VQA results by retrieving irrelevant or noisy facts \cite{marino2021krisp,li2022dynamic}. 

With the introduction of the pre-trained large-scale language models and their vast power in solving various tasks, recent approaches aimed to use the implicit knowledge accumulated in these models, such as GPT-3~\cite{brown2020language}. These models \cite{hu2023promptcap, shao2023prompting,yang2022empirical} performed many efforts to use LLMs and guide commonsense reasoning in VQA models. Despite their remarkable results, these methods lack powerful vision processing and suffer from hallucination.

In this work, we propose to train a supervised knowledge retrieval model using the provided meta-data in the KRVQA dataset \cite{cao2021knowledge}. Our retrieval method uses contrastive loss to maximize the similarity of a question embedding with the embedding of its supporting facts while minimizing its similarity to irrelevant facts. Moreover, we assume the corresponding scene graph (SG) of the image is given and utilize a similar retrieval module to extract the question-relevant visual information from the SG. In other words, we formulate both external knowledge and visual knowledge integration as retrieval tasks. Next, we integrate both external and visual retrieved knowledge to find the final answer through multiple hops of reasoning. 

The retrieved knowledge from KG\footnote{Since our external knowledge base is in graphical format, we use Knowledge Graph (KG) to mention our Knowledge Base (KB) interchangeably.} and SG  are integrated into both task-specific neural architectures (NN) and large language model(LLM) backbones. We analyze the strengths and weaknesses of the models regarding the question types and shed light on the remaining challenges in retrieval and integration steps for VQA task. In a nutshell, our contributions are as follows: 
\begin{itemize}
    \itemsep0em 
    \item We proposed a supervised model for retrieving relevant knowledge from the knowledge graph and scene graph conditioned on the question. Our results show the positive impact of empowering task-specific and LLM models with retrieved external and visual knowledge as well as their integration.
    \item We analyzed the impact of integrating external and visual knowledge in both task-specific neural architectures and LLMs for the VQA task. We found that LLM models outperform the NN model for KB-related questions which confirms the effectiveness of implicit knowledge in LLMs, however, they do not alleviate the need for external KB. 
    \item We investigated the strength of the task-specific and LLM models in multi-hop reasoning and performing both visual and KB reasoning. Our results show that though LLMs are stronger in 1-hop reasoning, they suffer in 2-hop reasoning in comparison with our fine-tuned NN model even in the presence of the supporting knowledge from both modalities. This highlights the importance of using a strong module for integrating both modalities and performing complex reasoning on top of that.
 
\end{itemize}
\section{Related Works}
We roughly divide recent VQA approaches into three groups as follows. The first group, plain VQA methods, focus on question and image integration approaches without including any extra information.  The second group aims to feed extra information extracted from external knowledge bases in VQA models. 
In the last group of SOTA VQA models, we review methods that rely on the power of the pre-trained Large Language Models (LLMs) as implicit sources of knowledge for KB-VQA tasks. 

\subsection{Plain VQA methods}
Similar to \cite{yu2018beyond}, MUTAN \cite{ben2017mutan} aims to use bilinear models for merging visual and textual information in VQA tasks using a multimodal Tucker tensor decomposition \cite{tucker1966some}. MUTAN parametrizes bilinear interactions between visual and textual representations to learn high-level associations between question meaning and visual concepts in the image while avoiding suffering from the curse of dimensionality. 

To focus on  both the visual and textual context of questions, the authors of MCAN \cite{yu2019deep} 
aim to design a ‘co-attention’ model to associate keywords in questions with key objects in images. 
They propose a deep Modular Co-Attention Network (MCAN) that is made by in-depth cascading of Modular Co-Attention (MCA) layers. Each  Modular Co-Attention layer (MCA layer) models the self-attention of questions and images, as well as the question-guided-attention of images jointly using a modular composition of two basic attention units.

The authors of \cite{ding2022mukea} suggest extracting and using knowledge triplets from training data, instead of feeding external knowledge bases. They propose to extract and represent multimodal knowledge by explicit triplets during training of VQA task. They discuss the complementary of the extracted triplets to the existing knowledge graphs and leave combining external knowledge bases open for the future.

\subsection{KB processing methods}

The MuCKO model \cite{zhu2020mucko} aims to answer the question by generating a semantic graph built on the dense captions of the image, a graph representing relevant knowledge triplets, and a spatial visual graph where nodes represent region features. It performs question-guided inter and intra-graph attention to answer the question iteratively. 

The authors of \cite{marino2021krisp} use Multimodal-BERT (MMBERT) \cite{khare2021mmbert} for their multimodal embedding and use Relational Graph Convolutional Network (RGCN) \cite{schlichtkrull2018modeling} to learn the representation of the concepts extracted from questions, Image, and KBs. To cover the knowledge required in OK-VQA, they construct a KB later.
Their model can improve marginally while suffering from huge memory and time demands. Moreover, since the supporting KB and the OKVQA dataset have been collected separately, this model is not analyzable. There are remaining questions on how VQA models can benefit from external KB while avoiding upcoming noise.



MAVEx is proposed by \cite{wu2022multi} aiming to leverage and integrate knowledge from visual (Google image search), textual (Wikipedia), and commonsense knowledge sources (ConceptNet). Their goal is to validate a set of candidate answers using answer-based knowledge retrieval.  

In \cite{guo2022unified}, the authors offer Unifer, a unified end-to-end retriever-reader framework for KB-VQA. To reduce noisy retrieval from external KBs, they aim to evaluate retrieved triplets based on their effect on final accuracy. 
They pay a huge cost for generating pseudo-labels to evaluate all knowledge triplets for each question. 

DMMGR \cite{li2022dynamic} represents the retrieved KB triplets in dynamic key-value memory format and converts visual content in a spatial graph representation.
It performs a question-knowledge guided attention over the spatial visual graph to find the required visual information to answer the question. 


\subsection{Large Language Models as Implicit Knowledge Sources}\label{LLM} 

Due to the revolutionary growth and power of large pre-trained models, the new trend in knowledge-based VQA aims to use large language-only pre-trained models like GPT-3 \cite{brown2020language} as implicit sources of knowledge, instead of using external sources of knowledge. Most often, these methods replace the visual data (image) with its corresponding tags and/or captions and then solve an unimodal text-only problem. 

The authors of \cite{gui2021kat} use a contrastive-learning-based module to retrieve knowledge from an external KB, and use GPT-3 to retrieve implicit knowledge by feeding image tags and captions as an input prompt. They integrate implicit and explicit knowledge and generate the final answer by integrating both implicit and explicit knowledge. The authors of \cite{yang2022empirical} propose PICa which relies on GPT-3 in joint retrieval of relevant knowledge and reasoning. They verbalize the visual signal by replacing the image with its caption and/or tags to provide more in-context information for the language model. 

The authors of \cite{lin2022revive} offer REVIVE targeting to integrate visual information, implicit knowledge, and explicit knowledge to generate the final answer. Similar to KAT, they have used a subset of Wikidata \cite{vrandevcic2014wikidata} as their KB. The authors of \cite{hu2023promptcap} propose Promptcap (Prompt-guided image Captioning) to control the visual entities in the generated caption using a textual query, and replace a general caption with a question-dependent caption. 



To avoid losing visual information, the authors of \cite{salaberria2023image} feed visual features alongside regional labels and captions to the transformer in their proposed Caption-Based Model(CBM).  
 Prophet's \cite{shao2023prompting} authors suggest including some candidate answers to enrich the prompt and improve final accuracy for LLM-based VQA.


It is worth mentioning that storing and retrieving knowledge implicitly in LLM parameters is prone to some errors happening during their knowledge storing and retrieving steps. These errors include but are not limited to 1. The knowledge hidden implicitly in LLM parameters is not necessarily complete, correct, or updated.  2. Correct memorization of knowledge during pre-training is not guaranteed, and 3. Retrieval of relevant knowledge is not guaranteed \cite{he2022rethinking}.

\section{The Proposed Method}
Our goal is to retrieve and integrate relevant knowledge triplets from the KGs to correctly answer a textual question related to a given image. 

There are two main challenges in this task, first, retrieving relevant knowledge from the external knowledge base, and second, integrating retrieved knowledge, the question, and the image to generate the final answer.  Regarding the first challenge, we propose a supervised model based on contrastive loss to maximize the similarity of the question embedding with the embedding of its supporting facts while minimizing their similarity to irrelevant facts.
Next, we replace our raw image with its corresponding scene graph and propose a similar contrastive-loss optimization method to retrieve the relevant visual information. We used the human-annotated scene graphs to represent the given visual data. For more information on the scene graph extraction and representation, one can check Appendix, section \ref{sec:sg}. 
Regarding integration and reasoning, we propose two frameworks based on task-specific neural architecture vs. using LLMs as the backbone to provide a comprehensive analysis of the impact of our retrieval method.  


\subsection{External Knowledge Retrieval}\label{sec:ret}

\begin{figure} [htb] 
    \centering
    \includegraphics[width=1\linewidth]{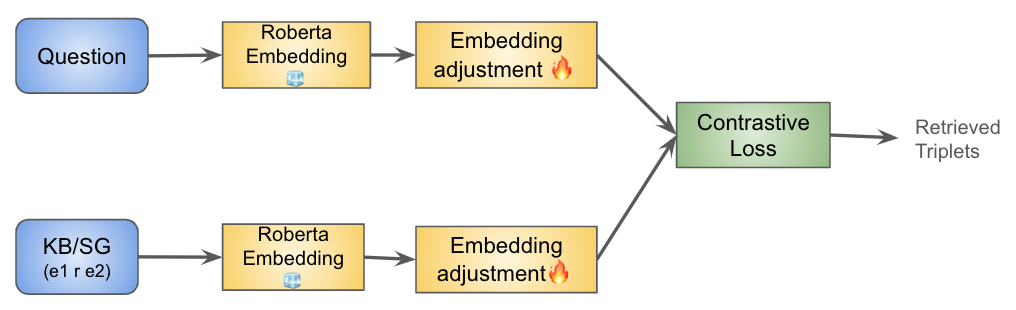}
    \caption{The Diagram of our supervised retrieval model. despite previous methods, we added some layers on top of frozen Roberta to adjust the Roberta embeddings to knowledge retrieval task using supporting knowledge presented in the ground-truth reason. }
    \label{fig:ret_arch}
\end{figure}

Since the most popular supporting knowledge bases provided for the VQA task are in graph format \cite{cao2021knowledge, marino2021krisp}, we propose a knowledge retrieval method for graphical knowledge-bases (KGs) which represents each fact in a triplet form, $(e_1, r_1, e_2)$ indicating first entity, relation, and second entity, respectively.
We propose an objective loss function to learn the representations of the triplets by contrasting positive and negative triplets. We convert each triplet $(e_1, r_1, e_2)$ to a sentence $e_1 r e_2$. One can use transformation-based graph encoders such as  TransE, QualE, etc~\cite{rossi2021knowledge}, or neural network-based models such as Relational Graph Convolutional Networks (RGCN)~\cite{schlichtkrull2018modeling} to initialize the triplet embeddings. However, the first group has been trained on a limited amount of KB data, and the RGCN family suffers from memory issues for data with a huge number of relations, and it only can learn embedding for entities and not for relations. Due to the availability of strong efficient pretrained text-encoders, we used Roberta \cite{liu2019roberta} for both plain text and triplet embedding initialization. We have used a feed-forward sub-layer on top of the Roberta representations to transform the embeddings to the same space. The diagram for our retrieval model is shown in Fig~\ref{fig:ret_arch}. 

 We use the following contrastive loss formulation, 
 \begin{align} \label{eq:CL}
    L_{Ret}(q,KB) = \sum_{t^+ \in pos} \sum_{t^- \in neg} [d(q, t^+) -d(q, t^-)],
\end{align}
 where q is the question, KB is the set of all KB triplets, $t^+$ is a set of triplets appearing in the supporting reason for the given question, and $t^-$ is a subset (the other triplets in the same batch) of all non-related triplets. $d(.,.)$ can be any distance measure, for which we use negative cosine similarity.

\begin{figure} [htb] 
    \centering
    \includegraphics[width=0.8\linewidth]{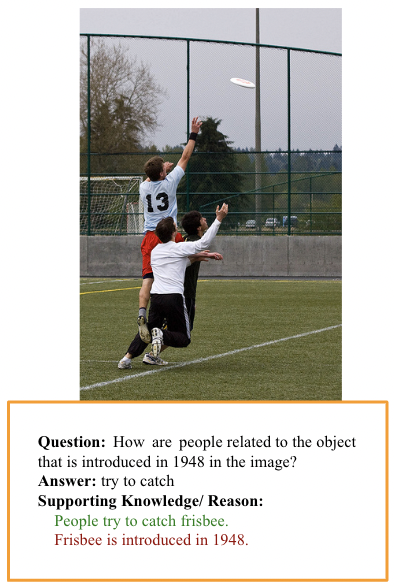}
    \caption{An example krvqa dataset. given image, question, answer, and Supporting reason. The green part of the reason is extracted from the image, and the red part from the external KB.}
    \label{fig:data}
\end{figure}

We use the same initialization method and architecture for both visual triplet retrieval from the scene-graph, and knowledge triplet retrieval from the external knowledge graph.
 It is worth mentioning that we have utilized the supportive reasons provided by KRVQA dataset for each question for supervising our retrieval module. We have used the image-related supporting facts (triplets) for supervising visual-retrieval training, and KB-related triplets for supervising KB retrieval training. An example of KRVQA dataset is shown in Fig \ref{fig:data}.
Since, the external KB includes $2339$ relations, $102343$ entities, and $225434$ facts, we first use keyword matching to find a smaller related subgraph before passing the graph to the retrieval step. We have extracted tags from the image, and keywords from the question, and extracted a subgraph by traversing two hops from each node, after finding KB nodes with matching keywords. It is noteworthy that this small subgraph is still very huge and has an order of $10K$ edges. 

\subsection{Integration and Reasoning Using Task-specific Neural Architectures}\label{sec:NN}

In our proposed task-specific architecture, we do multi-hop reasoning by integrating the retrieved KG and SG triplets in an repetitive manner using cross-attention units and generate the final answer as follows: 
\begin{align} \label{eq:reas}
& query_0 = question embedding, \\ \nonumber
    query_l = & Att(query_{l-1}, KB_{ret})+ \\
     & Att(query_{l-1}, SG_{ret})  \nonumber
\end{align}

\begin{figure} [htb] 
    \centering
    \includegraphics[width=1\linewidth]{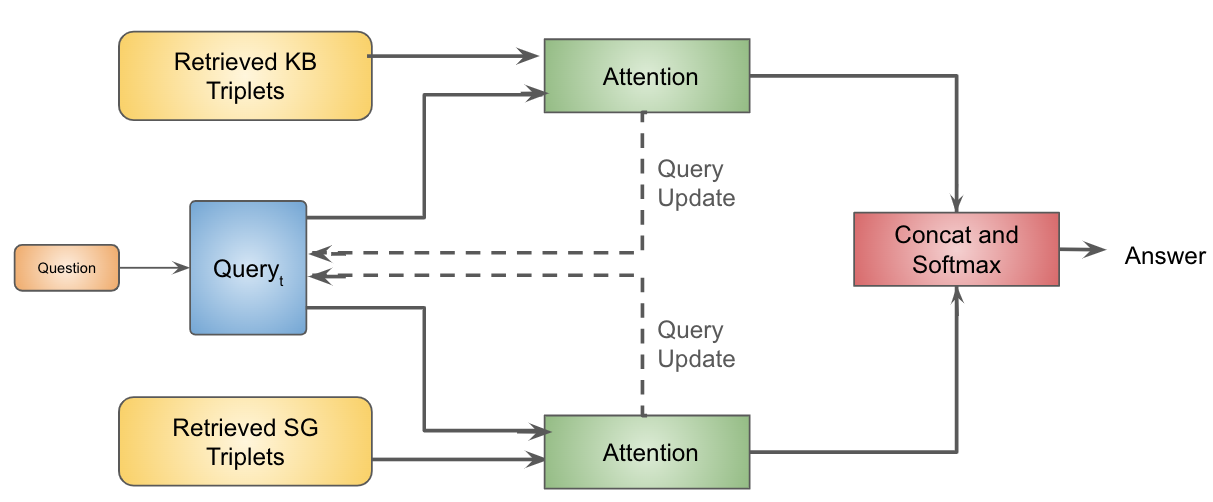}
    \caption{the architecture of our proposed task-specific neural architecture.  The attention modules iteratively update a query to solve multi-hop reasoning. The integration layer combines external and visual knowledge to provide the final output using a classifier layer. We initialize the query with question embedding and use KB and SG triplets as keys and values. The dashed arrows update the query iteratively using the output of attention-modules. }
    \label{fig:arch}
\end{figure}

 In other words, to handle several hops of reasoning, we update the cross-attention query (initialized by question embedding) by attending to retrieved KB and SG triplets in several layers. In the end, we combine the output of 2 cross-attention modules to perform the final VQA classification. The complete architecture of our task-specific neural model is shown in Fig \ref{fig:arch}.

\subsection{Integration and Reasoning using Large Language Models}\label{sec:LLM}

\begin{figure*} [htb] 
    \centering
    \includegraphics[width=0.99\linewidth]{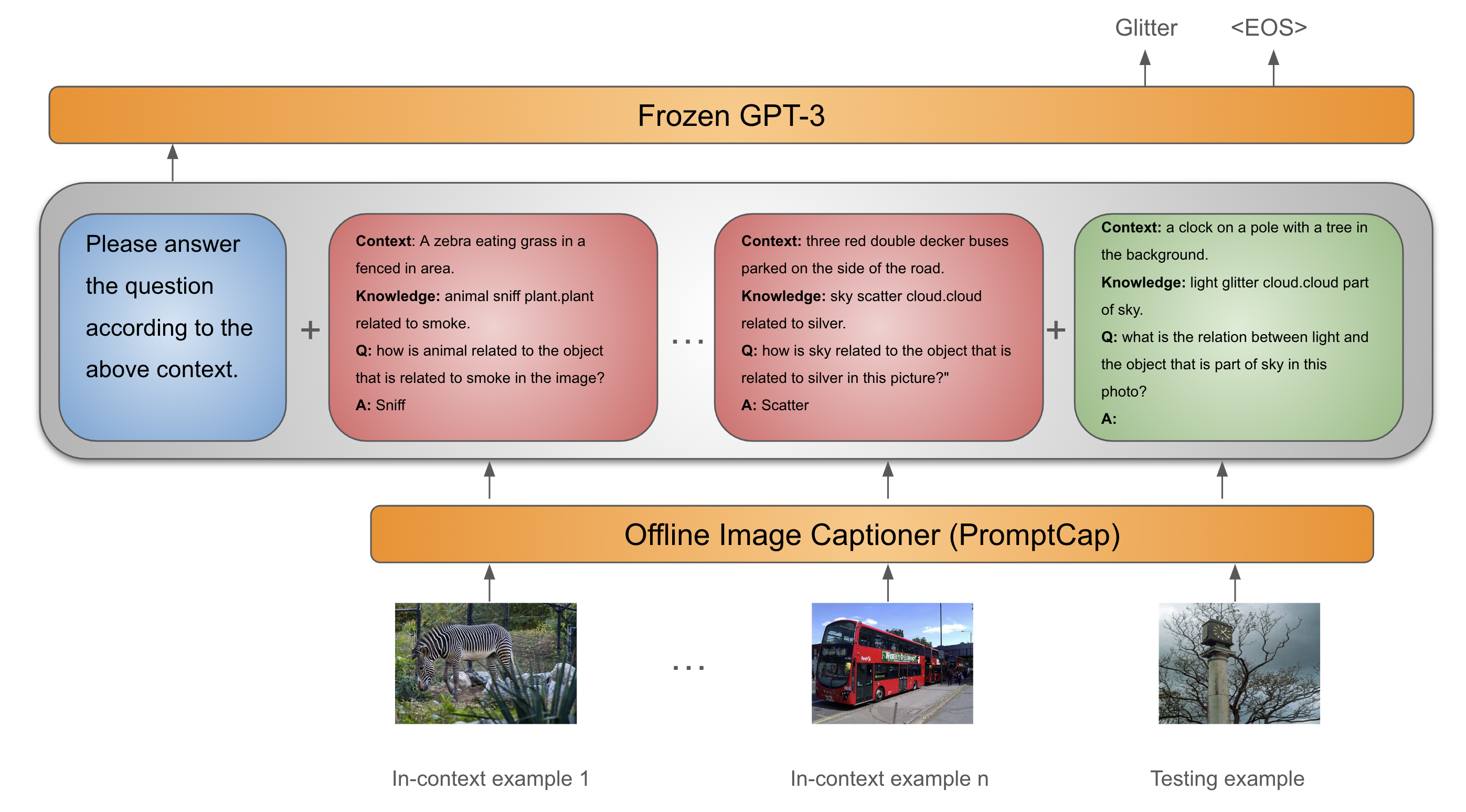}
    \caption{Architecture of our LLM model. The relevant knowledge is given as a part of the prompt }
    \label{fig:LLM}
\end{figure*}

Similar to \cite{hu2023promptcap,shao2023prompting, yang2022empirical}, we use a combination of the question and image caption as the input prompt for our LLM building block. 
For each test example, we find 32 training examples using maximum cosine similarity of image+question embeddings to form our few-shot prompts. We use CLIP\cite{radford2021learning} to embed the questions and images.  
To feed the supporting knowledge, we convert each triplet $(e_1, r_1, e_2)$ to a sentence $e_1 r e_2$, and inject it into the prompt as can be seen in the knowledge section as in Figure \ref{fig:LLM}. 
For the sake of fair comparison, we do not feed extra information such as image tags or candidate answers in our prompts. We use Promptcap captioner \cite{hu2023promptcap} to convert our images to text, and add these captions to the context section of the prompts.
Due to its significant accuracy of "davinci-002" for in-context learning, like \cite{hu2023promptcap, shao2023prompting} we use it as our LLM engine. 

\section{Experimental Results}
In the following sections, we will provide experimental results to show the impact of supervised knowledge retrieval and integration in both task-specific neural architectures and LLM frameworks for the VQA task. We aim to design experiments and provide results to answer the following questions:
1- Can we effectively augment task-specific and LLM VQA models by supervised retrieval of the relevant external and visual information? 2- How do task-specific and LLM-based models perform in multi-hop reasoning and integration of both external and visual sources of information? 
3. To what extent the implicit knowledge of LLMs can cover the required external knowledge for VQA task? 

\subsection{Datasets}\label{sec:dataset}

 \citeauthor{cao2021knowledge} propose knowledge-routed visual question reasoning (KRVQA) dataset that aims to prevent shortcut learning by generating non-crowdsourced question–answer pairs with their supporting reasons indicating which part of the reason is from visual, and which part is from external knowledge. It provides meta-data to analyze the strengths and weaknesses of the VQA models, for example, question type, supporting reasons divided into KB and visual parts, and number of reasoning hops to answer the questions. 
 
 KRVQA is to date the largest knowledge-based VQA dataset. It contains 32910 images and 157201 QA pairs. They are split into the train, validation, and test set at the ratio of 60\%, 20\%, and 20\%, respectively.  The questions in this dataset use external knowledge from FVQA dataset 225434 factual triplets~\cite{wang2017fvqa}. This external knowledge is being extracted from DBpedia\cite{auer2007dbpedia}, ConceptNet\cite{speer2017conceptnet}, and WebChild\cite{tandon2014webchild}. 
One of the current popular VQA datasets is OK-VQA~\cite{marino2019ok}. This dataset does not provide supporting facts/reasons for each question and lacks supporting metadata for our analysis goal. In this work, we use KRVQA dataset as its supporting meta-data helps to train our retrieval modules and provide a detailed analysis. 

\subsection{Implementation Details} \label{sec:imp}
 We train our model with PyTorch \cite{paszke2019pytorch} for all experiments using one GPU with 49GB of memory. For the contrastive loss, we treated all the other samples in a batch as negative samples. Our model is trained by AdamW \cite{loshchilov2018fixing} optimizer with 200 epochs, where the batch size is $256$ and the learning rate is set to $1*10^{-4}$. Cascading 4 layers of query-update and reasoning provides the best accuracy for integration and reasoning. We use top-1 accuracy as in \cite{cao2021knowledge} for a fair comparison. 
 
\subsection{Retrieval Result}
The accuracy of our retrieval module for both scene-graph and knowledge base is shown in Table \ref{tab:ret}. While top-1 retrieval accuracy for both SG and KB is around $60\%$, top-100 accuracy for both models is almost $100\%$. We can see that top-1 accuracy without training (using cosine similarity of text embedding) is less than half of the accuracy for the supervised retrieval.

\begin{table} [htb] 
    \centering
    \small
    \begin{tabular}{|c|c|c|} \hline
         & KB Retrieval & SG Retrieval\\ \hline
        Supervised & 59.10 & 60.00\\ \hline
        unsupervised & 17.5 & 27.12\\ \hline
    \end{tabular}
    \caption{Top-1 accuracy for our Retrieval on KRVQA dataset.}
    \label{tab:ret}
\end{table}

\subsection{Result and Discussion}\label{sec:acc}
Table~\ref{tab:top1} shows accuracy for VQA task using KRVQA dataset. 
\textbf{NN+($KB_{ret}$+$SG_{ret}$)} is our task-specific neural architecture while our retrieval model is used to retrieve the relevant part of the SG and KG (given the ground-truth scene graph and a relevant subgraph of the KG with preprocessing), as explained in section~\ref{sec:NN}. \textbf{LLM+($KB_{ret}$+$SG_{ret}$)} is our model with LLM backbone, enriched with relevant KB and SG triplets extracted using the method introduced in section~\ref{sec:ret}.

Our enriched LLM model improves accuracy (40.5\%) in comparison with Promptcap\cite{hu2023promptcap} (19.8\%) which is one of the most recent VQA frameworks using LLM. 

Our NN model has made more than 13\% improvements in comparison with the best NN model, DMMGR, which uses key-value memory networks to combine external knowledge in VQA tasks. This result is not a fair comparison though since we provide the ground-truth SG but our goal is to show the impact of supervised knowledge retrieval conditioned on access to the complete SG and KB. This result demonstrates the positive impact of training a supervised knowledge retrieval module for both external and visual knowledge besides a strong reasoning module to avoid noise propagation and enrich the model to perform better reasoning. The results show that enriching a base LLM model (Promptcap) by adding retrieved knowledge makes a significant improvement in accuracy, though still, a fine-tuned reasoning module leads to higher accuracy than LLM-based model.  
\begin{table} [htb] 
    \centering
    \small
    \begin{tabular}{L{5cm} | c} \hline
    Q-type \cite{cao2021knowledge}         & 8.12  \\
    LSTM \cite{cao2021knowledge}           & 8.81  \\
    FiLM \cite{perez2018film}              & 16.89 \\
    MFH \cite{yu2018beyond}                & 19.55 \\
    UpDown \cite{anderson2018bottom}       & 21.85 \\
    MCAN \cite{yu2019deep}                 & 22.23 \\
    Mucko \cite{zhu2020mucko}              & 24.00 \\
    KM-net \cite{cao2019explainable}       & 25.19 \\
    MUKEA  \cite{ding2022mukea}            & 27.38 \\
    DMMGR  \cite{li2022dynamic}            & 31.8  \\
    Promptcap \cite{hu2023promptcap}       & 19.8 \\ \hline \hline
    NN+($KB_{ret}$)                        & 22.28  \\ 
    NN+($SG_{ret}$)                        & 38.31 \\ 
    NN+($KB_{ret}$+$SG_{ret}$)             & 44.36  \\ \hline
    LLM+($KB_{ret}$)                       & 21.6  \\ 
    LLM+($SG_{ret}$)                       & 37.4 \\ 
    LLM+($KB_{ret}$+$SG_{ret}$)            & 40.50 \\ \hline
    \end{tabular}
    \caption{Top-1 accuracy percentage comparisons among different models on the KRVQA dataset.}
    \label{tab:top1}
\end{table}

As discussed in the appendix, we tried to replace the human-annotated scene graph with an approximated scene graph, but due to the low accuracy of current scene graph approximation methods, it did not lead to notable improvement and interpretable results. The remarkable improvement made by adding retrieved scene graph triplets demonstrates the importance of representing visual data in a high-level format instead of a low-level and abstract feature format.

\subsection{Question-type Analysis}\label{sec:anal}

The KRVQA dataset is generated using 7 different question-answer templates as shown in Table \ref{tab:krvqa2}.  We provided the accuracy of both our NN, and LLM-based models for each question type. $+KB_{ret}$ and $+SG_{ret}$ indicate feeding retrieved external and visual knowledge in the model respectively, while $+KB_{GT}$ and $+SG_{GT}$ indicate oracle setting and injecting ground-truth triplets in the models to find the upper-bound accuracy using our reasoning module.

 \begin{table*} [htb] 
     \centering
     \small
     \begin{tabular}{| c | L{10cm} | c | c|} \hline
         Qtype & Question Semantics & Answer & Reason\\ \hline
         0 & What is the relation of $<A>$ and $<B>$? & $<R>$ &   \\
         1 & What is $<A>$ $<R>$? & $<B>$ &  $(A,R,B)$\\
         2 & What $<R>$ $<B>$? & $<A>$ & \\ \hline
         3 & What is the relation of the object that $<A>$ $<R_1>$ and  $<C>$? & $<R_2>$ &  \\
         4 &  What is the relation of <A> and the object that $<R_2>$ $<C>$?& $<R_1>$ & $(A,R_1,B),(B,R_2,C)$ \\
         5 & What $<A>$ $<R_1>$ $<R_2>$? & $<C>$ & \\
         6 & What $<R_1>$ $<R_2>$ $<C>$? & $<A>$ & \\ \hline
     \end{tabular}
     \caption{KRVQA templates for the question-answer generation\cite{cao2021knowledge}}\par\bigskip
     \label{tab:krvqa2}
   \end{table*}

From results in Table \ref{tab:acc_q}, we can see that the proposed reasoning model is strong enough to find the final answer by combining the supporting facts, although even in the presence of ground-truth supporting facts/reasons, questions type 2, 3, and 5 are harder to answer using our NN model. Moreover, by comparing the accuracy using the retrieval model versus the upper limit of accuracy using ground truth triplets, our retrieval methods hurt more for question types 2,3, and 5. We conjecture that retrieving relevant facts is harder when the head entity is missing. 
As shown in Table \ref{tab:acc_hop}, for our retrieval-based model, KB-related questions are harder than non-KB-related models, although our retrieval accuracy for both SG and KB is similar. Moreover, the NN model has similar accuracy for 1-hop and 2-hop questions. 

Results show that question types 3 and 5 are harder for the LLM model even in the presence of ground-truth knowledge. Since for question types 3 and 5, the answer belongs to the second hop, it might make reasoning more complicated and prone to noise. Another interesting result is that almost for all one-hop questions LLM beats the NN model, though for 2-hop questions it performs worse than NN which confirms the weakness of LLM models in multi-hop reasoning.
In Table \ref{tab:acc_hop}, we can see that the LLM models work better than their corresponding NN models for KB-related questions which confirms the partial existence of implicit knowledge in LLMs. 

    \begin{table*}[htb]
    \centering
    \small
    \scalebox{0.99}{
    \begin{tabular}{ccccccccc} \hline
        Q-type& 0 & 1 & 2 & 3 & 4 & 5 & 6 & Overall\\ \hline
       NN+($KB_{ret}$+$SG_{ret}$)  & 63.34 & 56.76 & 23.63 & 14.75 & 54.55 & 10.28 & 54.29 & 89.31\\ \hline
       NN+($KB_{GT}$+$SG_{GT}$)    & 99.74 & 92.39 & 86.82 & 87.44 & 95.96 & 74.24 & 90.86 & 44.36\\ \hline \hline
       LLM+($KB_{ret}$+$SG_{ret}$) & 60.71 & 57.31 & 33.95 & 12.58 & 51.93 & 14.49 & 51.78 & 40.5\\ \hline
       LLM+($KB_{GT}$+$SG_{GT}$)   & 99.01 & 96.00 & 98.52 & 74.47 & 92.56 & 74.54 & 88.24 & 88.90 \\ \hline
     
    \end{tabular}
    }
    \caption{Accuracy for each question type.}
    \label{tab:acc_q}


   \centering
    \small
    \scalebox{0.99}{
    \begin{tabular}{ccccc} \hline
        Q-type& KB-related & No-KB-related & 1-hop & 2-hop \\ \hline
       NN+($KB_{ret}$+$SG_{ret}$)  & 34.63 & 43.33 & 37.99 & 38.83 \\ \hline
       NN+($KB_{GT}$+$SG_{GT}$) & 85.57 & 94.25 & 90.46 & 88.57 \\ \hline 
       LLM+($KB_{ret}$+$SG_{ret}$)  & 39.36 & 41.59 & 43.82 & 37.67 \\ \hline
       LLM+($KB_{GT}$+$SG_{GT}$)  & 91.19 & 91.15 & 98.09 & 80.68 \\ \hline \\
     
    \end{tabular}
    }
    \caption{Accuracy Analysis. }
    \label{tab:acc_hop}

    \centering
    \small
    \scalebox{1}{
    \begin{tabular}{cccccccc} \hline
        additional information & None & $KB_{ret}$ & $SG_{ret}$ & $KB_{ret}$+$SG_{ret}$ & $KB_{GT}$ & $SG_{GT}$ & $KB_{GT}$+$SG_{GT}$ \\ \hline
         NN  & 15.36 &  22.28 & 38.31 & 44.36 & 38.98 &67.06 &  89.48 \\ \hline
         LLM & 19.8 & 21.6 & 37.4 & 40.5 & 36.60 & 58.60 & 87.90 \\ \hline

    \end{tabular}
    }
    \caption{Ablation Study}
    \label{tab:ablation}
\end{table*}

\subsection{Ablation Study}\label{sec:abl}

We analyzed the impact of each source of knowledge (visual vs external) for our VQA task in Table \ref{tab:ablation}. Moreover,  We reported the effect of injecting retrieved knowledge vs the ground-truth knowledge which indicates the oracle setting and upper-bound for our models. 
The NN-None model only integrates question representation with image CLIP representation, while the LLM-None model does not use any knowledge in the prompts.

\textbf{NN model}: The results show that both proposed KB and SG retrieval models make improvements in accuracy. Moreover, we can see that our proposed multi-hop reasoning is strong enough to make reasoning and find the final answer from the ground-truth supporting facts (accuracy equal to 89.48\%). The huge gap between the None and $SG_{ret}$ model shows the importance of providing a more high-level (object-level) representation of the visual data, despite a low-level abstract representation of an image or a general caption. 

\textbf{LLM model}: The results for LLM-based model support the previous findings, though similar to the results in the previous section (Table \ref{tab:top1}), we can see that LLM models are moderately less accurate in comparison with the task-specific fine-tuned model (NN), which highlights the impact of a strong knowledge integration and reasoning module. Supporting the findings for NN model, the huge improvement resulting from adding visual knowledge retrieval demonstrates that a caption does not adequately represent the visual data, and we need an alternative approach to feed the visual data.

\section{Conclusion and Future Work}

We proposed a retrieval and reasoning framework for knowledge-based visual question answering. We trained a contrastive-loss model to retrieve supporting facts from external knowledge as well as from the scene graph. We used the KRVQA dataset, the largest existing VQA dataset, to train both retriever and reasoning modules.

Our results show the positive impact of empowering task-specific and LLM models with retrieved external and visual knowledge.
We investigated the strength of the task-specific and LLM models in multi-hop reasoning, as well as integrating visual and KB reasoning. Our results show that though LLMs are stronger than the fine-tuned NN model in 1-hop reasoning, they suffer in 2-hop reasoning even in the presence of the supporting knowledge from both modalities. This highlights the importance of using a strong module for the integration of both sources of information and performing complex reasoning.
We found that LLM models outperform the NN model for KB-related questions which confirms the effectiveness of implicit knowledge in LLMs, however, they do not alleviate the need for external KB. 

\section{Limitations}
\begin{itemize}
\itemsep0em 
    \item We trained and executed retrieval and reasoning parts separately. We will investigate the impact of dynamic fact retrieval interchained with reasoning in the future. 
    \item Due to the huge size of our external knowledge base, we retrieve the relevant knowledge for each question only one time before starting the reasoning step. we will replace this static retrieval module with a dynamic retrieval module which refines the retrieved knowledge after each step of reasoning.
    \item we used human-annotated scene graphs provided in the Visual Genome dataset, as approximated scene graphs do not generate acceptable accuracy. In the future, we have to replace VG scene graphs with another automatically generated high-level representation of the image.
    \item Some commonly-used VQA datasets like OK-VQA do not provide supporting facts/reasons for each question and lack supporting metadata to analyze the strengths and weaknesses of the reasoning models. Therefore, we did not provide results on this dataset. In the future, we try to modify our models and perform experimental results on more datasets.
\end{itemize}
\bibliographystyle{acl_natbib}
\bibliography{anthology}

\section{Appendix}

\subsection{Scene Graph Computation} \label{sec:sg}
We treated both the knowledge base and visual signal in graph format. We proposed a supervised retrieval method to retrieve relevant triplets from the external knowledge graph and scene graph. Since the images in the KRVQA dataset are taken from the Visual Genome dataset \cite{krishna2017visual}, we have used the scene graphs provided by the VG dataset in our method. These scene graphs on average include 36 objects and 22 relationships per image.

 To have a more realistic setting, we made efforts to provide results with approximated scene graphs as well. Given an image I, we used Faster-RCNN \cite{ren2015faster} to identify a set of objects $O=\{o_i\}^K_{i=1}, (K=36)$, where each object $o_i$ is associated with a visual feature vector $v_i \in R^{d_v}, (d_v=2048)$, and a corresponding label. The final scene graph is shown as $G_V=(V^V,E^V)$ over objects $O$, where $V^V={v^V_i }^K_{i=1}$ is the node set and each node $v^V_i$ corresponds to a detected object $o_i$. We have trained our scene graph generation method using the training data introduced in \cite{xu2017scene} (VG150). VG150 includes the most frequent 150 object categories and 50 predicates of the VG dataset. We use 70\% of the images for training and the remaining 30\% for testing. We generated scene graphs using both VS3 \cite{zhang2023learning}  which used GLIP \cite{li2022grounded} as a building block for object and feature recognition, as well as the scene graph benchmark proposed by \cite{tang2020unbiased}. Unfortunately, scene graph detection quality is low, and error propagation from scene graph generation hampers the retrieval accuracy from 60\% to around 19\% (as shown in table \ref{tab:ret}). Therefore, we use Visual Genome provided scene graphs in our work to avoid error propagation from the scene graph to the next steps of the model and have a better analysis of the strengths and limitations of our retrieval and reasoning model.

\end{document}